# Classification of Motor Imagery EEG Signals by Using a Divergence Based Convolutional Neural Network

**ZümrayDokur, Tamer Ölmez**

Istanbul Technical University, Department of Electronics and Communication Engineering, Istanbul, Turkey
*corresponding author: dokur@itu.edu.tr*

**ABSTRACT:** Deep neural networks (DNNs) are observed to be successful in pattern classification. However, high classification performances of DNNs are related to their large training sets. Unfortunately, in the literature, the datasets used to classify motor imagery (MI) electroencephalogram (EEG) signals contain a small number of samples. To achieve high performances with small-sized datasets, most of the studies have employed a transformation such as common spatial patterns (CSP) before the classification process. However, CSP is dependent on subjects and introduces computational load in real-time applications. It is observed in the literature that the augmentation process is not applied for increasing the classification performance of EEG signals. In this study, we have investigated the effect of the augmentation process on the classification performance of MI EEG signals instead of using a preceding transformation such as the CSP, and we have demonstrated that by resulting in high success rates for the classification of MI EEGs, the augmentation process is able to compete with the CSP. In addition to the augmentation process, we modified the DNN structure to increase the classification performance, to decrease the number of nodes in the structure, and to be used with less number of hyper parameters. A minimum distance network (MDN) following the last layer of the convolutional neural network (CNN) was used as the classifier instead of a fully connected neural network (FCNN). By augmenting the EEG dataset and focusing solely on CNN's training, the training algorithm of the proposed structure is strengthened without applying any transformation. We tested these improvements on brain-computer interface (BCI) competitions 2005 and 2008 databases with two and four classes, and the high impact of the augmentation on the average performances are demonstrated.

## 1. INTRODUCTION

The brain computer interface systems enable communication between computers and humans by analyzing the EEG data. Actually, the purpose of BCI studies is to convert EEG signals to computer commands. In recent years, many studies have been carried out on BCI applications. Among the developing methods, nowadays, we observe that a significant amount is on the classification of motor imagery EEG signals. In MI EEG classification, the subject imagines the movement of his/her hands, feet and tongue, and the acquired EEG signals are analyzed and converted into computer commands. Motor imagery can be thought as a mental preview of a movement without any physical output. It is widely accepted that, similar areas on brain are active during mental imagery or during execution of a movement.

The EEGs are biological signals which do not have distinctive waveforms such as electrocardiogram or heart sounds; the shape of the EEG is in the form of noise. When working with MI EEG signals, one should know that (*i*) these signals are of low spatial resolution, (*ii*) there are variations between subjects and sessions, (*iii*) it is not so easy to classify many different classes with high success rates, and (*iv*) collecting data from inexperienced subjects is a serious problem. The spatial resolution is directly related with the number of electrodes placed over the head. Although higher resolutions can be obtained by using more and more electrodes, the computational load of the BCI system will be increased. Besides, using many electrodes makes the BCI experiments to be unpractical and uncomfortable for the subjects. Imagination of the same movement greatly varies among the subjects, even between two sessions at different times for the same subject. Unfortunately, the number of commands is limited with the motor imageries; there are at most four commands (classes) in the current literature. Another concern is that, the same classifier may give high performance for one subject while showing poor performance for another one. Hence, there is a need for the training of the subjects; the subjects should be made ready for the experiments by practicing the MI tasks before the final data collection phase.

In the classification of MI EEG, it is important to determine the computer commands not only correctly, but also quickly in terms of system efficiency. However, we observe that researchers' main focus is only on increasing the classification success rates, and most of them prefer benchmark datasets [1, 2] to compare their methods with the studies in the literature. As a result, time and memory consuming complicated feature extraction methods and classifiers had been developed to increase the success rates [3–40]. However, the complexity of the algorithms prevents to run the classification algorithm on mobile/portable embedded systems. In addition, it is very difficult to determine the optimum features in complicated structures. Moreover, the researchers need to spend too much effort to determine subject-based optimum features. It is also important to use a dataset with a large amount of data to evaluate the classification performances. However, the benchmark datasets [1, 2] in the literature are of small sizes, and it is observed that researchers mostly prefer only one of these datasets to demonstrate the classification performances of their methods [3–14, 19–24, 28–35].

In this study, we have investigated whether the training of DNN can be strengthened by using the existing small-sized MI EEG datasets or not. The goal is to strengthen the DNN's training and thus achieve high classification performances for MI EEG signals using a small-sized network. In the literature, the classification performances obtained for MI EEG signals are not high enough. As the number of classes increases, the success rates for MI EEG signals decrease drastically. In this respect, there is a need to increase the classification performance of MI EEG signals especially for four classes. It is observed in the literature that, complicated algorithms used to increase the performances make the system even more impractical. For this reason, in this study, in order to increase the classification performance for small-sized MI EEG datasets, we have focused on strengthening the training of DNNs. We suggest two approaches to strengthen the training of DNNs: (*i*) Using a novel augmentation process on small-sized MI EEG datasets, and (*ii*) using a DNN structure which is proposed in this study.

### 1.1 Classification of MI EEG signals in the literature

In the early studies, traditional neural networks have been used to classify the MI EEG signals [28–38]. In this case, the CSP, Fourier transform and discrete wavelet transform are observed to be the most employed feature extraction methods in classifying the MI EEGs. The traditional neural networks, such as, LDA (linear discriminant analysis), SVM (support vector machine) and MLP (multi layer perceptron) have been preferred as a classifier in the classification of MI EEG signals. Table 1 shows the classification results obtained with traditional neural networks by using the benchmark datasets [1, 2]. As can be observed from the table, high success rates could not be achieved with traditional neural network topologies, and the average performance decreases when the number of classes increases. The classification performance is very dependent on the correct determination of the features. Therefore, the researchers worked heavily on determining the optimum features for each subject.

In our previous studies [15–17], we have focused on to increase the classification performances of the MI EEG signals by using the CSP together with the traditional neural networks, because CSP is frequently used to increase the classification performances of the MI EEG signals. We made an effort to determine parameters of the CSP optimally for each subject, and focused on creating new models derived from CSP. But, classification performances were not as high as those obtained in the studies using the DNNs.

In the meantime, the DNN structures have been continuously developing and achieving high performances in classification problems [41–46]. In recent years, it is observed that DNNs are widely used in the classification of MI EEG signals [3–14, 25–27, 47–52]. It is observed that success rates obtained with DNNs are higher than those obtained with traditional neural networks. In addition, one of the advantages of DNNs is that there is no need for extra effort to determine the features. CNN automatically extracts the features from the dataset during the training.

Table 1: The classification of MI EEG signals obtained with the traditional neural networks.

| methods | accuracy % | number of classes | transformation + feature extraction | classifier | dataset |
|---|---|---|---|---|---|
| Wang's method [28] | 77.2 | 4 | CSP + variance | MLP | BCI III-IIIa |
| Aljalal's method [29] | 80.2 | 2 | Wavelet + statistical, entropy, energy features | MLP | BCI III-IVa |
| Mirnaziri's method [30] | 61.7 | 4 | CSP + variance | MLP | BCI IV-IIa |
| Silva's method [31] | 67.8 | 2 | Linear Predictive Coding | MLP | BCI IV-IIb |
| Wang's method [36] | 81.2 | 2 | FD-CSP + variance | SVM | BCI IV-IIb |
| Alansari's method [32] | 83.8 | 2 | Wavelet | SVM | BCI IV-IIb |
| Mishuhina's method [37] | 89.8 | 4 | RCSP – FWR | LDA | BCI III-IIIa |
| Behri's method [33] | 89.4<br>94.5<br>67.4 | 2 | Wavelet | SVM<br>K-NN<br>MLP | BCI III-IVa |
| Zhang's method [34] | 84 | 2 | CSP + variance | SVM | BCI III-IVa |
| Li's method [35] | 68.6 | 3 | FBCSP+ variance | SVM | BCI IV-IIa |
| Molla's method [38] | 92.2<br>91.3<br>91.2 | 2 | CSP + subband features | SVM<br>LDA<br>K-NN | BCI III-IVa |

Some of studies in deep learning attempt to classify the MI EEG signals into three classes [47, 48] and some of them use data not related to the BCI competitions [49–51]. It is observed that BCI Competition III datasets IIIa and IVa [1], and BCI Competition IV datasets IIa and IIb [2] are generally preferred to evaluate the classification performances as reference datasets [3–38, 47, 48, 52]. BCI Competition III datasets IIIa and IVa have MI EEG records from three and five subjects, respectively. BCI Competition IV datasets IIa and IIb each has MI EEG records from nine subjects. Deep neural networks need big dataset. However, the number of MI EEG recordings in the datasets is still not high enough. To achieve high performances with small-sized datasets, it is observed in the literature that transformations such as CSP [3, 4, 6, 7, 11, 14–17, 25–27, 39, 40], fast Fourier transform (FFT) [5], short-time Fourier transform (STFT) [9, 10], continuous wavelet transform (CWT) [12] and others [8, 13] had been employed before the CNN to increase the classification performance for the MI EEG signals. In [52], it is observed that DNN's success on MI EEGs decreases when transformation is not applied to the raw input data. Another observation is that, in MI EEG classification tasks, the achievements for the two-class problems are considerably higher than those for the four-class problems. Therefore, transformation process seems to be almost a necessity to achieve high performances with four-class problems. In the classification of MI EEGs, CSP is the most preferred method to transform the raw input data. Transformation of the signal to a different domain by using the CSP was basically considered as a preprocessing step. The CSP-like transformations act as a guide in the determination of the feature planes of DNNs. However, CSP is dependent on subjects and introduces computational load in real-time applications. Also, it is necessary to optimize the parameters used in the CSP in order to increase the classification performance. Especially, the $m$ parameter of the CSP affects the performance of the transformation and also the overall classification performance [15−17].

### 1.2 Motivation and contribution

Classification performance for four-class MI EEG signals is still not at a sufficiently high level. Fortunately, it is observed that DNNs produce high classification performances in almost all areas. Besides their benefits, the DNNs have the following two major drawbacks among the others: (*i*) DNNs need big datasets, and (*ii*) determination of even a coarse structure of DNN may take days. We have noticed that by solving some problems encountered in DNNs, classification performance for the MI EEG signals can be improved.

Unfortunately, the number of MI EEG recordings in BCI Competition datasets is also not high enough. A solution for increasing the classification performance with small datasets is to use transfer learning [12]; in the transfer learning, the huge network pre-trained using a different and probably big data set is retrained by the BCI MI EEG dataset. In [12], high successes are achieved, but the DNN really has excessive number of nodes. In another solution, researchers mostly preferred using preprocessing or transformation stage before the CNN structure to provide high classification performances with small datasets [3−14, 25−27]. The CSP [3, 4, 6, 7, 11, 14, 25−27, 39, 40], FFT [5], STFT [9, 10] or CWT [12] are preferred for the preprocessing (or transformation) stage. In those studies, CNNs investigate the features in the input space determined by the preprocessing. However, these stages introduce considerable amount of computational load to the decision making system. Because, since, the classification performance is very dependent on the correct determination of the parameters of the preprocessing stage, in addition to a successful training of the DNN, the researchers need to accomplish a successful preprocessing in order to determine the optimum parameters for each subject.

As mentioned above, one of the major problems is the determination of a coarse structure for the DNN. The smaller the DNN structure, the easier it is to search for a coarse model. As the structure of the network departs from complexity, a stronger control is achieved over the network, which in turn leads to effective training of the network. At the same time, in real-time applications, large networks present some problems both in the training of the networks and in generating classification results (*i.e.* test responses). Small CNN structures are preferred in order to generate the classification responses fast and to avoid memory problems [3, 4, 6–9, 52].

Leaving the existing studies aside, we propose, for the first time, (*i*) to enrich the training set of the MI EEG signals with augmentation, and (*ii*) to use a DNN structure with fewer hyper parameters in order to strengthen the training of the DNNs. This approach will increase the classification performance of MI ECG signals without the need to use a preprocessing stage. In this study, the augmentation process is applied to EEG data for the first time in the literature, and the training sets are synthetically enriched by using the novel augmentation method. In case of not using a preprocessing stage, the positive effect of the augmentation process on the classification performance is demonstrated by using the paired t-test in the *Computer Simulations* section. In general, DNNs consist of two parts: a CNN (feature extractor), and a fully connected neural network (FCNN) or another neural network such as LVQ (linear vector quantization) and SVM, among others. In some cases, the DNN contains only the CNN or the FCNN. In general, the FCNN owns the majority of the nodes of DNN. In this study, the proposed DNN is comprised of the CNN and a minimum distance network as classifier instead of FCNN. Therefore, the proposed method focuses solely on the training of CNN, as the minimum distance network is not trained.

In this context, three contributions have been provided to the literature: (*i*) the best classification accuracies according to the literature have been obtained for four different datasets by strengthening the training of DNNs, (*ii*) since the main building blocks (CNN + FCNN) of the DNN can be trained individually, hyper-parameters are quickly set, and features in the CNN are determined better, and (*iii*) since the classification results (*i.e.* test responses) have been obtained in a short time by using a small-sized DNN, the proposed method is suitable for use in real time applications.

## 2. METHODOLOGY

In this section an augmentation process which is proposed for the first time in the literature, and the structure and training of the proposed CNN to be used in the classification of MI EEG signals will be explained. In this study, two classification models are investigated: (*ii*) proposed DNN, and (*ii*) transformation + proposed DNN.

## 2.1 Augmentation process for motor imagery EEG signals

The purpose of creating a training set with the augmentation process is to direct the CNN to find the right features without being affected by redundant information on the data. In this context, four processes which are explained below are performed to augment the MI EEG signals. These four processes were applied to both BCI Competitions 2005 and 2008 datasets. 80% of all data is allocated for the training set and the remaining 20% for the test set. The augmentation process is done only for the training set. The original EEG signals are also included in the augmented dataset.

In the first process of the augmentation, the mean values of all EEG signals in all channels are made equal to zero. In the second process, the EEG signals in all the channels are amplified by multiplying the amplitudes with the same value selected randomly between 0.2 and 5.0. In the third process, the polarity of the EEG signals in all channels is inverted with a probability of 0.5. In other words, EEG signals are multiplied by −1 or +1 with a probability of 0.5. In the fourth process, the EEG in the epoch is rotated around a single value selected randomly between (−*Epoch_Size*/2) and (+*Epoch_Size*/2). In the fifth and the last process, random noise with a standard deviation of 0.01 is added to the EEG signals in all channels. The pseudo code equivalents of the above processes which are all given in Algorithm 1 is applied to each epoch of the training set.

The five processes in Algorithm 1 are sequentially applied to each EEG record in the training set, and the set of five operations is repeated nine times. After these processes, the training dataset is increased ten times with the original EEG signal also included in the augmented set.

In MI experiments, it was observed that there were variations in measured power at special frequencies which are called $\mu$ (8−12 Hz) and $\beta$ (18–26 Hz) bands [15]. After the augmentation processes (setting the mean to zero, amplification by random amount, polarity inversion, rotation along time dimension, and random noise injection), spectral content of motor imagery signals is not lost in the new epochs. In the *Computer Simulations* section, the positive effect of the augmentation on the classification performance will be shown with the paired t-test.

The augmentation algorithm is run once and a new training set is created. This new training set is not only created for training the proposed CNN, it can also be employed for the training of any conventional DNN. The augmentation algorithm is independent of the CNN, and it is only associated with enlarging the data within the training set.

## 2.2 Structure of the proposed convolutional neural networks

In general, a traditional DNN structure comprises of two cascaded units: a convolutional neural network which in effect plays the role of the "feature extractor (FE)", and a fully connected neural network as the "classifier". The CNN structure is a combination of the convolutional layer, pooling layer, and rectified linear units (ReLU) layer.

**Algorithm 1** The pseudo code equivalent of augmentation algorithm.

$E$: number of channels, $N$: epoch size, $X$ [0:$E$, 0:$N$]: epoch matrix

**1- Setting the mean value to zero**

**for** $i$ = 0 to $E$ **do**
  set the mean value of data in $X$ [ $i$, : ] to zero
**end for**
$X_1$[ : , : ] ← $X$[ : , : ]

**2- Amplification by random amount**

$ra$ ← randomly select one number in range [0.2, 5.0]

$X_2$[ : , : ] ← $X_1$[ : , : ] × $ra$

**3- Polarity inversion**
  $rp$ ← randomly select one value in the set {−1 , +1}
  $X_3$[: , :] ← $X_2$[ : , : ] × $rp$

**4- Rotation along the time dimension**

$rr$ ← randomly select one integer number in range [−$es$/2, +$es$/2]

**for** $i$ = 0 to $E$ **do**
  rotate data in $X_3$[ $i$ , : ] by $rr$
**end for**
$X_4$[ : , : ] ← $X_3$[ : , : ]

**5- Random noise injection**

$RM$[ : , :] ← randomly generate a matrix of ($nc$×$es$) by using normal distribution N($mean$ = 0, $sd$ = 0.01)

**for** $i$ = 0 to $E$ **do**
  **for** $j$ = 0 to $N$ **do**
    $X_5$[ $i$ , $j$ ] = $X_4$[ $i$ , $j$ ] + $RM$[$i$ , $j$ ]
  **end for**
**end for**

**return** $X_5$

The convolutional layer is the major processing layer in the CNN architecture which has a set of kernels (learnable filters) with small receptive fields. These filters are convolved with the input signal resulting in a 1D activation map for each filter. The pooling layer in the CNN performs a nonlinear down-sampling operation, and in the meantime, contributes to the translation invariance of the network. By inserting a pooling layer between the CNN's convolutional layers, the input size of the CNN is progressively reduced. Among several nonlinear functions, max-pooling is the most preferred pooling operation in machine learning applications. The need for using the pooling layers varies according to the problem and there is no general methodology. In the literature, pooling process is widely used in CNNs, but, this layer does not necessarily need to be used after every convolutional layer. In this study, a pooling layer is used between all the successive convolutional layers. With an activation function of $f(x)$=max(0, $x$), the ReLU layer sets the negative values in the activation maps to zero, thus performs a nonlinear operation in the decision making system. Moreover, the batch normalization and drop-out functions are used between each layer to further strengthen the training. While the convolutional neural network generally represents the features, the fully connected neural network represents the classifier. It is difficult to determine the hyper-parameters of CNN and FCNN, and to train both structures simultaneously. When these two parts of DNN are employed, the number of hyper-parameters of DNN increases drastically, some of which are: (*i*) number of convolutional layers, (*ii*) number of feature planes in convolutional layers, (*iii*) filter size, (*iv*) number of hidden layers in FCNN, and (*v*) number of nodes in the hidden layers. It may take days to determine the optimum values of these parameters. If the DNN consists of only the CNN, the number of hyper-parameters of DNN decreases to: (*i*) number of convolutional layers, (*ii*) number of feature planes in convolutional layers, and (*iii*) filter size, *etc*.

In this study, the proposed DNN is comprised of the CNN, and a minimum distance network instead of the FCNN as the classifier. The proposed network provided in Fig. 1 is called DivFE which is the acronym for the *Divergence-based Feature Extractor*. The FCNN and its training phase introduces a lot of problems, some of which are: (*i*) failing to converge to the optimum solution, (*ii*) high computing load in real time applications, (*iii*) high memory requirements in portable systems (such as FPGA or embedded Linux cards), and (*iv*) determination of the hyper-parameters such as the number of hidden layers and the number of nodes in these hidden layers.

After several convolutional layers, a MDN is used to determine the class (label) of the input signal. The weights of the nodes of the MDN do not change during the training. These weights are set to the values in the rows (or columns) of the Walsh matrix. The output of the proposed feature extractor is fed to the input of the MDN. The node of the MDN that is the closest to the FE's output signal determines the class of the input signal. MDN determines the label of the class of the input signal by using the equations below:

$$\boldsymbol{D}_k = \sum_{j=1}^{M} (\boldsymbol{OFE}_j - \boldsymbol{RW}_{k,j})^2 \qquad \boldsymbol{D}_i = \min_k(\boldsymbol{D}_k) \tag{1}$$

where $M$ represents the dimension of the Walsh matrix (in this study $M$ is equal to 16), $\boldsymbol{OFE}_j$ represents the $j$th output of the feature extractor (and also outputs of the flatten layer in Fig. 1), $\boldsymbol{RW}_{k,j}$ represents the $j$th element of the Walsh vector belonging to the $k$th class, index $i$ represents the label of the class which is the decision of the MDN.

### 2.3 Training the convolutional neural networks

Training both the convolutional and fully connected layers of deep networks simultaneously for each input and output pair makes it difficult to determine the weights of the network optimally.

Because, the risk of being caught by the local optimum points increases as the number of nodes of the network takes high values. In this study with the developed training strategy, it is possible to train the convolutional and fully connected layers individually. If both are desired to be included in the network, first the convolutional layer and then the fully connected layer can be trained sequentially. If they are trained separately, the features are more likely to be accurately determined with this training strategy. As a result of the achievements in the extracted features, the fully connected neural layers can be eliminated. Therefore, we preferred to use only a simple minimum distance classifier instead of the fully connected layer.

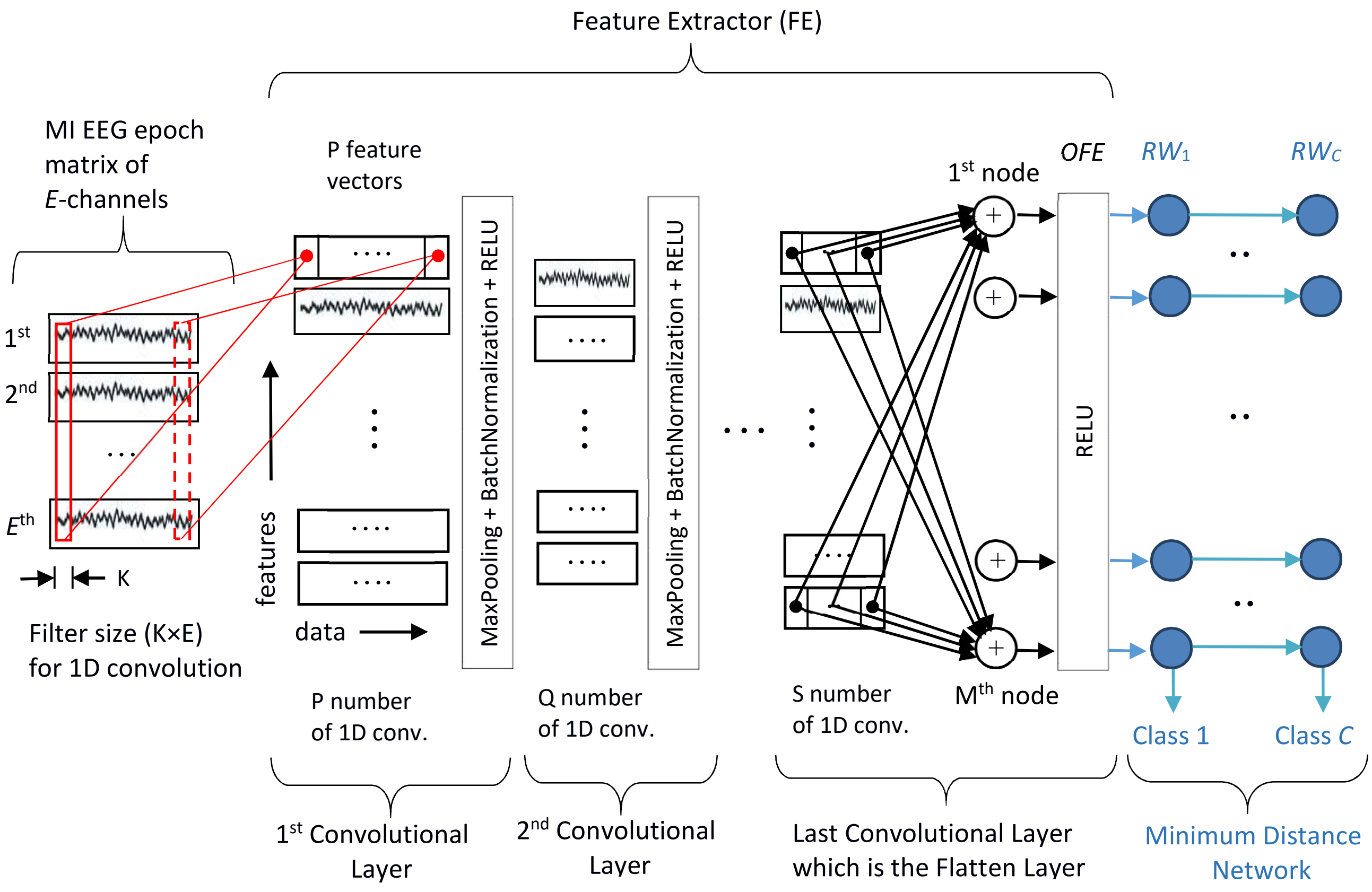

**Fig. 1** The structure of the DivFE. OFE represents the flatten layer. The size of the OFE vector is equal to the size of Walsh vectors. The structure of the Feature Extractor is not different from the CNN known in the literature. In this study a minimum distance network (MDN) is proposed instead of the fully connected neural network. $\boldsymbol{RW}_k$ are the nodes of the MDN constituted with the rows of Walsh matrix. The weights of $\boldsymbol{RW}_k$ are not updated in the training.

In the developed training strategy, it is aimed to achieve a high divergence value for the feature space searched by the CNN layer. In classification problems, the divergence value computed over a feature set, points to the effectiveness of the features in terms of being intra-class representative and inter-class discriminative. In this respect, a class separability definition is made as given in the following equation:

$$\text{divergence value} = \operatorname{tr}(\boldsymbol{SW}^{-1} \cdot \boldsymbol{SB})$$
$$\boldsymbol{SW} = \boldsymbol{SW}_1 + \boldsymbol{SW}_2 + \cdots + \boldsymbol{SW}_C \quad (2)$$

Here, tr is the trace operation, $\boldsymbol{SW}_k$ is the covariance matrix of the $k$th class, and the sum of the covariance matrices of all the $C$ classes is denoted by $\boldsymbol{SW}$ which is also called the within-class scatter matrix of the distribution. The $\boldsymbol{SB}$ is the between-class scatter matrix, and it is the covariance of the mean vectors of the classes. According to Eq. (2), the ratios of larger between-class scatters to smaller within-class scatters lead to high divergence values to be obtained. A high divergence value signifies that a favorable distribution of the feature vectors has been attained.

In this regard, to distribute the mean vectors (centers) of the classes at the farthest positions in the feature space with equal distances from each other, the Walsh vectors showed up to meet these requirements. By associating each row (or column) of the Walsh matrix with a class center, given an input vector, the FE is trained to output the specific row (or column) of the Walsh matrix which represents the class of that input vector. With this training strategy, selecting the output vectors of the FE as the rows of the Walsh matrix increases the distances between the class centers, thus increases the divergence value which is an indication of better features.

To realize the ReLU function in the FE, 0 and 1 values had to be used in place of −V and V in the Walsh matrices, respectively (as shown for $\boldsymbol{W}_8$ in Eq. (3)). In Eq.(3), original Walsh matrices for two- and four-dimensional spaces and a modified Walsh matrix for an eight-dimensional space are given.

$$\boldsymbol{W}_2 = \begin{bmatrix} +V & +V \\ +V & -V \end{bmatrix} \quad \boldsymbol{W}_4 = \begin{bmatrix} +V & +V & +V & +V \\ +V & -V & +V & -V \\ +V & +V & -V & -V \\ +V & -V & -V & +V \end{bmatrix} \quad \boldsymbol{W}_8 = \begin{bmatrix} 1 & 1 & 1 & 1 & 1 & 1 & 1 & 1 \\ 1 & 0 & 1 & 0 & 1 & 0 & 1 & 0 \\ 1 & 1 & 0 & 0 & 1 & 1 & 0 & 0 \\ 1 & 0 & 0 & 1 & 1 & 0 & 0 & 1 \\ 1 & 1 & 1 & 1 & 0 & 0 & 0 & 0 \\ 1 & 0 & 1 & 0 & 0 & 1 & 0 & 1 \\ 1 & 1 & 0 & 0 & 0 & 0 & 1 & 1 \\ 1 & 0 & 0 & 1 & 0 & 1 & 1 & 0 \end{bmatrix} \quad (3)$$

In the modified Walsh matrix, it can easily be noticed that the Hamming distance between any two rows or columns is equal to the half of the matrix rank value (rank is equal to the dimension of output vectors). Therefore, the nodes of the MDN are chosen from the rows (or columns) of the modified Walsh matrix. In this study, Walsh matrix rank was chosen as 16 for both the two- and four-class problems. The first two- and four rows in the Walsh matrix are reserved to represent the class centers (nodes) of MDN for two- and four-class problems, respectively. Increasing the matrix rank also increases the Hamming distances between the class centers, contributing to the class separability criterion. On the other hand, over increasing the rank should be avoided as it causes prolonged training phases. Fig. 2 shows the training algorithm of the DivFE.

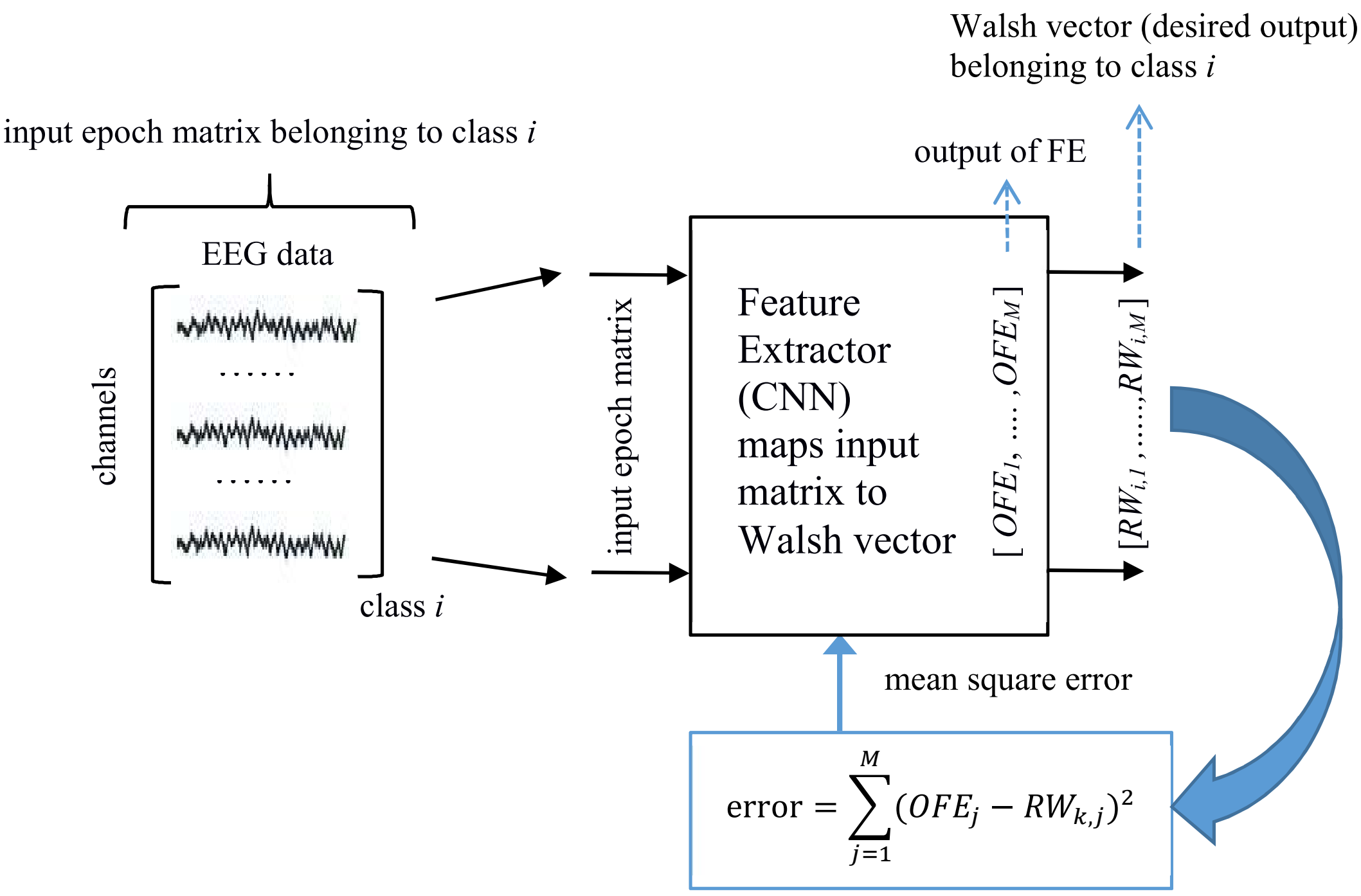

(a) Training process without CSP transformation

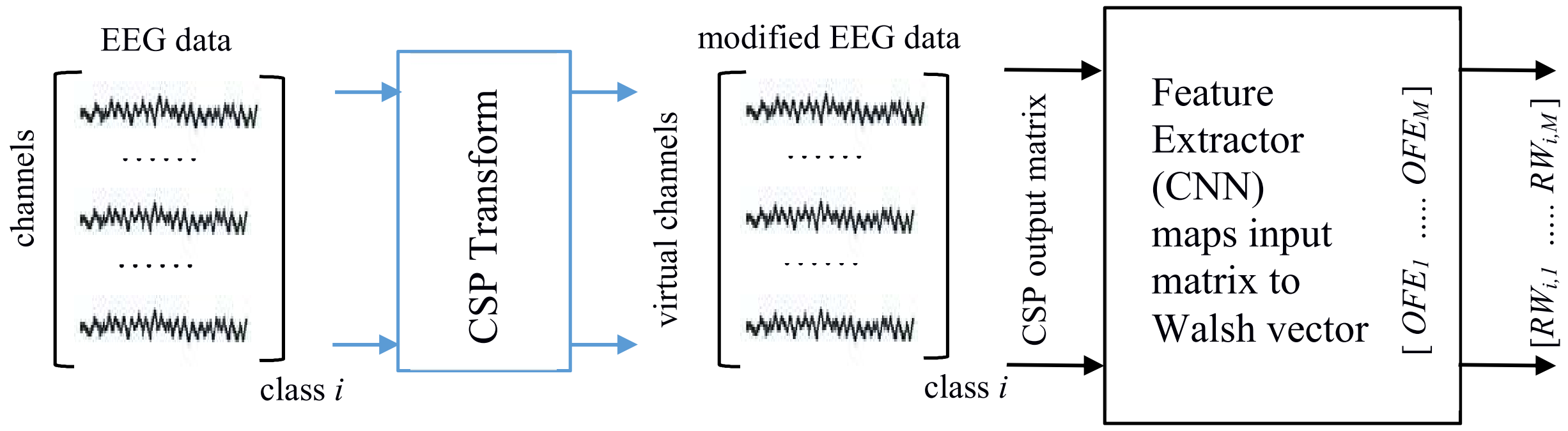

(b) Training process with CSP transformation

Fig. 2 Training process of DivFE. $\boldsymbol{OFE}_j$ represents the $j$th output of the feature extractor (and also the outputs of the flatten layer in Fig.1), and $\boldsymbol{RW}_{i,j}$ represents the $j$th element of Walsh vector belonging to the $i$th class. In this study $M$ is equal to 16. As an example, for two classes: $\boldsymbol{RW}_1$=[1,0,1,0,1,0,1,0, 1,0,1,0,1,0,1,0] and $\boldsymbol{RW}_2$=[1,1,0,0,1,1,0,0,1,1,0,0,1,1,0,0].

Walsh functions are preferred due to the following four properties: ($i$) It is known that in any feature space to satisfy the divergence criterion at its best, both the class centers should locate distantly from each other as much as possible, and within-class scatterings should be small. Since the nodes of the MDN hold the centers of the classes, by representing each class center by a row (or column) of the Walsh matrix, the centers of the classes distribute over the feature space at the farthest possible distances. This approach increases the divergence value. ($ii$) In general, increasing the number of features (the number of weights in the nodes of MDN) improves the classification performance. In the proposed method, the dimension of the feature space can be set to high values by simply increasing the rank of Walsh matrix. This approach will provide a better class distribution in a multi-dimensional space. (*iii*) In deep learning, the best performing features and the classifier weights are both searched simultaneously during the training phase. Since convergence problems may happen with such dense networks, in the proposed method it is suggested to train the CNN (feature extractor) and the classifier individually. In this study, MDN is preferred as a classifier. By copying the content of the Walsh matrix to the nodes of the MDN, the classifier is made ready to be used without iterative training. Hence during the training only the CNN is trained to map the input vectors to Walsh vectors, and the MDN is only run during the testing phase. Because the FCNN was not used, the training algorithm was focused only on the features, so eventually, a decrease in the within-class scatterings was achieved. (*iv*) Transfer learning (the use of pre-trained networks) has also become a popular choice for the applications with deep learning. A pre-trained network designed with a Walsh matrix of rank greater than the number of classes, can easily be expanded to learn the distribution of the new additional classes. In this study as the rank of Walsh matrix was 16 for both the two- and four-class problems studied, if requested, the pre-trained network can be retrained for the additional 14 or 12 new classes, respectively. In the proposed study the dimension of the feature space is not kept too large in order to obtain the classification results quickly, and to make the network learn the dataset quickly.

### 2.4 Dataset preparation and validation process

Four different experiments will be covered in order to examine the effects of the augmentation and transformation on the classification of MI EEG signals: (*i*) no augmentation + no CSP, (*ii*) no augmentation + CSP, (*iii*) augmentation + no CSP, and (*iv*) augmentation + CSP. These experiments are applied to four different datasets. First of all, 80% of each dataset is allocated for the training and the remaining 20% for the test. If augmentation process will be used in the experiments, a new training set is formed by augmenting the epochs in the existing training set. If the CSP will be used in the experiments, the weights of CSP transform are determined by using only the epochs in the training set. And then, CSP transform is applied to all the epochs in the training and test sets. Thus, new training and test sets are obtained.

The differences between the training and validation processes, and the structures of a basic DNN and DivFE are pointed out in Algorithm 2. 10% of the training set is selected as the validation set. After all the training data is used for once for the training of FE (it is known as one iteration), accuracy and mean squared error loss values are calculated for all data in the validation set. In the validation phase, each input data (epoch) of the validation set is given to the FE, and FE generates an output vector for each input data. This output data (OFE) is presented to the MDN. The node of the MDN that is the closest to the OFE determines the label of the class for the input data as decision. After the assigned Walsh vectors for each input data are compared with the vectors generated by the MDN, mean accuracy and loss values are calculated over all the data in the validation set.

**Algorithm 2:** Differences between the training and validation processes, and the structures of a basic DNN and DivFE for two classes

**Basic DNN:** Feature Extractor (FE convolutional neural network) + Classifier (FCNN)

**1- One iteration of training process for two classes**

**for** each input data in {training set} **do**
    **run** the network (FE + FCNN) and obtain the *network output* $O = [O_1, O_2]$
    **define** *desired output*: $[0, 1]^T$ for the input data of the first class
    $[1, 0]^T$ for the input data of the second class
    **update** weights of network (FE+ FCNN) by using $\sum(\textit{network output} - \textit{desired output})^2$
**end for**

**2- Validation/Test process for two classes**

**for** each input data in {validation/test set} **do**
    **run** the network (FE + FCNN) and obtain the *network output* $O = [O_1, O_2]$
    **obtain** the decision ($i$) of basic DNN by calculating the below equation

$$O_i = \max(O_1, O_2)$$

**end for**
**calculate** accuracy by comparing decision of basic DNN for each input data with the class label of the input

**DivFE** shown in Fig. 1: Feature Extractor (FE convolutional neural network) + Classifier (MDN)

**1- One iteration of training process for two classes**

**for** each input data in {training set} **do**
    **run** the network (FE) and obtain the *network output OFE*
    **define** *desired output*: $RW_1 = [1,0,1,0,1,0,1,0,1,0,1,0,1,0,1,0]^T$ for the input data of the first class
    $RW_2 = [1,1,0,0,1,1,0,0,1,1,0,0,1,1,0,0]^T$ for the input data of the second class
    **update** weights of network (FE) by using $\sum(\textit{network output} - \textit{desired output})^2$
**end for**

**2- Validation/Test process for two classes**

**for** each input data in {validation/test set} **do**
    **run** the network (FE) and obtain the *network output* $OFE = [OFE_1, \dots, OFE_{16}]$
    **obtain** the decision ($i$) of DivFE by calculating the below equation for $RW_1$ and $RW_2$

$$\text{MDN:} \quad D_k = \sum_{j=1}^{M} (OFE_j - RW_{k,j})^2 \qquad D_i = \min_k(D_k)$$

**end for**
**calculate** accuracy by comparing decision of DivFE for each input data with the class label of the input

While determining a coarse structure for the FE, mean squared error loss values and accuracies for both the training and validation sets are examined to avoid underfitting and overfitting. During this stage, hyper-parameters (filter size, number of layers, number of features in the layers, *etc.*) of the FE are determined by trial and error, as done in the literature. After the coarse structure is determined, FE's training is started for long iterations. The training is terminated considering both the iteration number and the loss value for the validation set. And then, the accuracy for the test set is calculated. For $N$ randomly generated training and test sets, the FE is trained $N$ times and the average of $N$ accuracies for the test sets is calculated to achieve statistically significant success.

## 3. COMPUTER SIMULATIONS

All the analyses were realized using Python codes running on Ubuntu Linux workstation which had 32 core CPUs of 2.7 GHz with GeForce GTX1080 Ti Graphics card.

In the classification of MI signals, the proposed modifications in the structure and training of CNNs are tested on two commonly used databases in the literature, BCI Competitions III (2005) and IV (2008) databases [1, 2], where 80% of whole data collected from the subjects is allocated for the training and the remaining 20% for the test. In this study, a separate CNN was determined for each subject. In this section, all the tables show the averages of 30 experiments with random training and testing sets. Tables 3, 5, 11 and 13 present the success of each class in terms of positive predictive value (ppv) and negative predictive value (npv).

The MI signals of two (left hand and right hand) and four classes (left hand, right hand, foot and tongue) are examined separately for each BCI competition database. Therefore, the proposed contributions are experimented on four different datasets. Two different models are compared: (*i*) no transformation is applied, and (*ii*) transformation is applied in the classification. In the case of 'no transformation' (Fig. 2a), EEG signals are supplied directly to the proposed FE. In the other case (Fig. 2b), both the bandpass filters and CSP method are used together as the transformation. Each channel is analyzed within five different frequency bands (6−12, 12−18, 18−24, 24−30, 30−36 Hz). Therefore, inputs as many as (the number of channels)×(filter banks) are presented to the CSP. The number of outputs (virtual channels) of the CSP is related to the $m$ parameter. This parameter defines the input dimension of the proposed DivFE structure.

In the tables showing FE structures the first, second and third values represent the number of input nodes of FE, size of the convolution filter and the number of feature planes, respectively. In this study the optimum $m$ parameters of the CSPs are investigated to increase the classification performance. Therefore, in the FE structures trained with each subject's own data, the numbers of input nodes of the networks are different in the tables.

Recent studies, using DNNs, on the classification of MI EEG signals are summarized in Table 2. The four rows at the bottom of the table show the classification results obtained by the proposed methods in this study. It is observed that Lu's study [5] achieved a classification accuracy of 84%. However, this accuracy is obtained for a dataset with feedback, in which the subjects get feedback via the screen in order to produce correct motor imagery signals. On the other hand, since our goal was to investigate the effects of augmentation process on the classification performance of MI EEGs, we did not prefer to use feedback.

In table 1, it was observed that SVM and LDA, instead of the MLP, are generally used in the classification of MI EEG signals. There is no feature extraction process in this study; instead, the CNN is preferred to perform the feature extraction task. However, when a traditional neural network is used, it is needed to determine a feature extraction method. It is observed that the variances of the channels are frequently used as the features after the CSP transformation [15]. In order to compare the results obtained by using basic neural networks with the results of the proposed CNN, the MI EEGs in the four datasets are classified with MLP classifiers. Table 3 presents performance results obtained using CSP as the transformation, variances of channels as the feature extraction, and MLP as the classifier. In literature [28–38], there are several studies in the literature for the classification of MI EEGs by using the traditional neural networks (such as MLP, SVM and LDA, among others.) and different feature extraction methods on the same datasets. The classification performances of MI EEG signals obtained with the traditional neural networks were shown in Table 1. As can be observed from the Table 1, high success rates could not be achieved with traditional neural network topologies.

### 3.1 Classification of MI EEG signals in BCI Competition III

In this subsection, MI signals of two and four classes will be examined separately. In this respect, the two-class BCI Competition III (2005) dataset IVa and four-class BCI Competition III (2005) dataset IIIa will be introduced, and results of the experiments involving transformation and no transformation applied to these datasets will be presented.

#### 3.1.1 Classification of two-class MI EEG signals

BCI Competition III (2005) dataset IVa contains epochs belonging to two classes collected from five subjects. Each epoch contains 68-channel EEG signals. 80% of all data collected from the subjects is allocated for the training set and 20% for the test set. There are a total of 280 epochs for each subject, and the number of epochs in the training set is increased by ten times after the augmentation process. Therefore, the number of epochs of the training and test sets for each subject are $10\times280\times0.8$ and $280\times0.2$, respectively.

Table 4 shows the classification results of the MI EEG signals processed without using a transformation. The subjects are numbered to associate them with their corresponding FE structures which are uniquely designed for each subject to generate the classification results. These numbers are also related with the numbers explained in the description of the BCI Competition III (2005). Table 4 also shows the comparisons between the proposed DivFE and the studies in the literature in terms of classification accuracies. Table 5 presents the number of input nodes, the size of the convolution filter, and the number of feature planes (number of filters, *i.e.*, depth of the convolution layer) at each layer in the FE structures designed for each subject.

**Table 2** Classification results of DNNs for MI EEG signals.

| studies | database -dataset | number of classes | accuracy % | transformation | number of subjects |
|---|---|---|---|---|---|
| Yang's study[3] | IV-IIa | 4 | 69.27 | Filter + CSP | 9 |
| Sakhavi's study [4] | IV-IIa | 4 | 70.6 | Filter + CSP | 9 |
| Lu's study [5] | IV-IIb | 2 | 84 | FFT | 9 |
| Sakhavi's study [6] | IV-IIa | 4 | 74.46 | Filter + CSP | 9 |
| Abbas's study[7] | IV-IIa | 4 | 70.7 | Filter + CSP | 9 |
| Kar's study[50] | IV-IIa | 4 | 70.5 | no transformation | 9 |
| Wu's study[8] | IV-IIb | 2 | 80.6 | Filter bank | 9 |
| Dai's study[9] | IV-IIb | 2 | 78.2 | STFT | 9 |
| Tabar's study[10] | IV-IIb | 2 | 77.6 | STFT | 9 |
| Thang's study[11] | III-IIIa | 4 | 91.9 | CSP | 3 |
| Chaudhary's study[12] | III-IVa | 2 | **99.3** | CWT | 5 |
| Zhao's study[13] | IV-IIa | 4 | 75 | 3D-EEG | 9 |
| Zhang's study [14] | IV-IIb | 2 | 82 | CSP+bispectrum | 9 |
| Deng's | III-IIIa | 4 | 85.3 | FBCSP | 3 |
| study [25] | IV-IIa | 4 | 78.9 | FBCSP | 9 |
| Olivas-Padilla's | IV-IIa | 4 | 78.4 | DFBCSP | 9 |
| study[26] | | | for monolithic network | | |
| Liu's study [27] | IV-IIa | 4 | 76.86 | CSP | 9 |
| Soman's study [39] | III-IIb | 2 | 76.3 | CSP | 9 |
| NTS-A-III/IVa | III-IVa | 2 | 96.2 | no transformation | 5 |
| TS-A-III/IVa | | | 98.5 | CSP | |
| NTS-A-III/IIIa | III-IIIa | 4 | **96.5** | no transformation | 3 |
| TS-A-III/IIIa | | | 95.8 | CSP | |
| NTS-A-IV/IIa | IV-IIa | 4 | **79.3** | no transformation | 9 |
| TS-A-IV/IIa | | | 79.1 | CSP | |
| NTS-A-IV/IIb | IV-IIb | 2 | **88.6** | no transformation | 9 |
| TS-A-IV/IIb | | | 85.1 | CSP | |

Fig. 3 shows the loss values for the training and validation sets with respect to the iteration number for subject S1 in Table 4. Training is terminated based on the iteration number and the loss value for the validation set.

Table 6 shows the classification results of the MI EEG signals processed by using transformation (bandpass filters + CSP). As mentioned in the above paragraph, the subjects are numbered to associate them with their corresponding FE structures which are uniquely designed for each subject to generate the classification results. Table 6 also shows the comparisons between the proposed DivFE and the studies in the literature in terms of the classification performances. Table 7 presents number of input nodes, filter size, and number of feature planes at each layer in the FE structures which generate the classification results in Table 6.

Table 3: Comparison of the proposed DNNs with multi-layer perceptron in the classification of MI EEG signals.

| methods | accuracy % | number of classes | transformation + feature extraction | classifier | dataset |
|---|---|---|---|---|---|
| realized in this study | 61.2 | 4 | CSP + variance | MLP | BCI III-IIIa |
| realized in this study | 75.5 | 2 | CSP + variance | MLP | BCI III-IVa |
| realized in this study | 61.2 | 4 | CSP + variance | MLP | BCI IV-IIa |
| realized in this study | 63.7 | 2 | CSP + variance | MLP | BCI IV-IIb |
| proposed method | **96.5** | 4 | (no transform) DivFE | MDN | BCI III-IIIa |
| | 95.8 | 4 | CSP + DivFE | MDN | BCI III-IIIa |
| | 96.2 | 2 | (no transform) DivFE | MDN | BCI III-IVa |
| | **98.5** | 2 | CSP + DivFE | MDN | BCI III-IVa |
| | **79.3** | 4 | (no transform) DivFE | MDN | BCI IV-IIa |
| | 79.1 | 4 | CSP + DivFE | MDN | BCI IV-IIa |
| | **88.6** | 2 | (no transform) DivFE | MDN | BCI IV-IIb |
| | 85.1 | 2 | CSP + DivFE | MDN | BCI IV-IIb |

**Table 4** Classification results for the two-class MI EEG signals without using transformation.

| | % positive and negative predictive values for each class<br>% accuracies obtained for each subject | | | | | |
|---|---|---|---|---|---|---|
| | **aa (S1)** | **al (S2)** | **av (S3)** | **aw (S4)** | **ay (S5)** | Mean % Accuracy (Kappa) |
| ppv for class 1 by DivFE | 100 | 100 | 74.4 | 100 | 96.4 | |
| npv for class 2 by DivFE | 96.2 | 100 | 100 | 96.4 | 100 | |
| average accuracy of DivFE | 98.1 | 100 | 87.2 | 98.2 | 98.2 | 96.3 (0.926) |
| Wang's study [18] | 95.5 | 100 | 80.6 | 100 | 97.6 | 94.7 |
| Ang's study [19] | 93 | 98 | 94 | 94 | 89 | 93.6 |
| Higashi's study [20] | 92.2 | 99.2 | 78.0 | 99.2 | 95.0 | 92.7 |

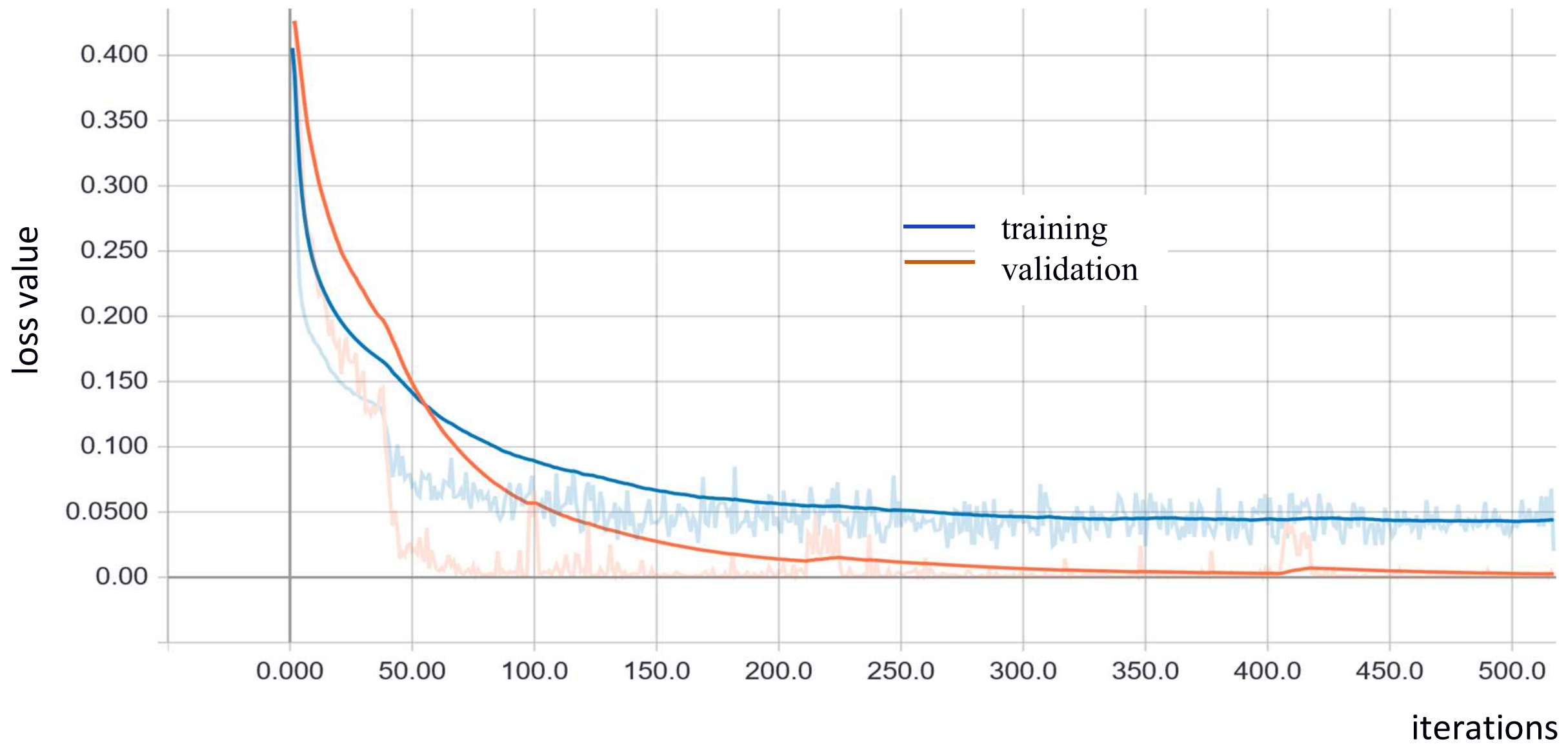

Fig. 3. Plot of loss values for training and validation sets according to iterations for subject S1 in Table 4.

**Table 5** The FE structures which generate the classification results in Table 4.

| FE layers | #input nodes, filter size, #feature planes in FE layers | | | | |
|---|---|---|---|---|---|
| | **S1** | **S2** | **S3** | **S4** | **S5** |
| **Layer 1** | 68, 9, 40 | 68, 9, 40 | 68, 9, 40 | 68, 9, 40 | 68, 9, 40 |
| **Layer 2** | 40, 9, 40 | 40, 9, 40 | 40, 9, 40 | 40, 9, 40 | 40, 9, 40 |
| **Layer 3** | 40, 9, 40 | 40, 9, 40 | 40, 9, 40 | 40, 9, 40 | 40, 9, 40 |
| **Layer 4** | 40, 9, 40 | 40, 9, 40 | 40, 9, 40 | 40, 9, 40 | 40, 9, 40 |
| **Layer 5** | 40, 9, 40 | 40, 9, 40 | 40, 9, 40 | 40, 9, 40 | 40, 9, 40 |
| **Layer 6** | 40, 8, 16 | 40, 8, 16 | 40, 8, 16 | 40, 8, 16 | 40, 8, 16 |

### 3.1.2 Classification of four-class MI EEG signals

BCI Competition III (2005) dataset IIIa contains epochs belonging to four classes collected from three subjects. Each epoch contains EEG signals of 43 channels. 80% of whole data collected from subjects is allocated for the training and the remaining 20% for the test. There are a total of 240 epochs for each subject, and the number of epochs in the training set is increased by ten times after the augmentation process. Therefore, the number of epochs in the training and test sets for each subject are $10 \times 240 \times 0.8$ and $240 \times 0.2$, respectively.

Table 8 shows the classification results of the MI EEG signals processed without using transformation, and presents the comparisons between the proposed DivFE and the studies in the literature in terms of classification performances. Table 9 presents the FE structures specifically designed for each subject to generate the classification results in Table 8.

**Table 6** Classification results for the two-class MI EEG signals by using transformation

| | % positive and negative predictive values for each class<br>% accuracies obtained for each subject | | | | | Mean % Accuracy (Kappa) |
|---|---|---|---|---|---|---|
| | **aa (S1)** | **al (S2)** | **av (S3)** | **aw (S4)** | **ay (S5)** | |
| ppv for class 1 by DivFE | 100 | 100 | 100 | 100 | 100 | |
| npv for class 2 by DivFE | 100 | 100 | 87.4 | 100 | 100 | |
| average accuracy of DivFE | 100 | 100 | 92.8 | 100 | 100 | 98.5 (0.97) |
| Wang's study [18] | 95.5 | 100 | 80.6 | 100 | 97.6 | 94.7 |
| Ang's study [19] | 93 | 98 | 94 | 94 | 89 | 93.6 |
| Higashi's study [20] | 92.2 | 99.2 | 78.0 | 99.2 | 95.0 | 92.7 |

**Table 7** The FE structures which generate the classification results in Table 6.

| | #input nodes, filter size, #feature planes in FE layers | | | | |
|---|---|---|---|---|---|
| **FE layers** | **S1** | **S2** | **S3** | **S4** | **S5** |
| **Layer 1** | 2, 7, 40 | 2, 7, 40 | 8, 11, 60 | 10, 7, 40 | 2, 7, 40 |
| **Layer 2** | 40, 7, 40 | 40, 7, 40 | 60, 11, 60 | 40, 7, 40 | 40, 7, 40 |
| **Layer 3** | 40, 7, 40 | 40, 7, 40 | 60, 11, 60 | 40, 7, 40 | 40, 7, 40 |
| **Layer 4** | 40, 7, 40 | 40, 7, 40 | 60, 11, 60 | 40, 7, 40 | 40, 7, 40 |
| **Layer 5** | 40, 16, 16 | 40, 16, 16 | 60, 11, 60 | 40, 16, 16 | 40, 16, 16 |
| **Layer 6** | | | 60, 8, 16 | | |

**Table 8**Classification results for the four-class MI EEG signals without using transformation

| | % accuracies obtained for each subject | | | |
|---|---|---|---|---|
| | **k3b (S1)** | **k6b (S2)** | **l1b (S3)** | Mean % Accuracy (Kappa) |
| DivFE for class 1 | 100 | 90 | 100 | |
| DivFE for class 2 | 93.2 | 93 | 94 | |
| DivFE for class 3 | 100 | 94 | 93.6 | |
| DivFE for class 4 | 100 | 100 | 100 | |
| average accuracy of DivFE | 98.3 | 94.3 | 96.9 | 96.5 (0.953) |
| Guan's study [21] | 86.6 | 81.6 | 85 | 84.4 |
| FBCSP method in [15] | 92.4 | 70.7 | 80.2 | 81.1 |
| DFBCSP method in[15] | 93.7 | 73.2 | 84.7 | 83.8 |

Table 10 shows the classification results of the MI EEG signals processed by using transformation (bandpass filters + CSP), and presents the comparisons between the proposed DivFE and the studies in the literature in terms of the classification accuracies. Table 11 presents the FE structures specifically designed for each subject to generate the classification results in Table 10.

**Table 9** The FE structures which generate the classification results in Table 8.

| | #input nodes, filter size, #feature planes in FE layers | | |
|---|---|---|---|
| **FE layers** | **S1** | **S2** | **S3** |
| **Layer 1** | 43, 9, 40 | 43, 9, 40 | 43, 9, 40 |
| **Layer 2** | 40, 9, 40 | 40, 9, 40 | 40, 9, 40 |
| **Layer 3** | 40, 9, 40 | 40, 9, 40 | 40, 9, 40 |
| **Layer 4** | 40, 9, 40 | 40, 9, 40 | 40, 9, 40 |
| **Layer 5** | 40, 9, 40 | 40, 9, 40 | 40, 9, 40 |
| **Layer 6** | 40, 8, 16 | 40, 8, 16 | 40, 8, 16 |

**Table 10** Classification results for the four-class MI EEG signals by using transformation.

| | % accuracies obtained for each subject | | | |
|---|---|---|---|---|
| | **k3b (S1)** | **k6b (S2)** | **l1b (S3)** | |
| DivFE for class 1 | 91.8 | 93 | 100 | |
| DivFE for class 2 | 98.3 | 95.2 | 97 | |
| DivFE for class 3 | 98.3 | 100 | 95 | Mean % Accuracy (Kappa) |
| DivFE for class 4 | 85.2 | 100 | 96 | |
| average accuracy of DivFE | 93.4 | 97.1 | 97.0 | 95.8 (0.944) |
| Guan's method [21] | 86.6 | 81.6 | 85 | 84.4 |
| FBCSP method in [35] | 92.4 | 70.7 | 80.2 | 81.1 |
| DFBCSP method in [35] | 93.7 | 73.2 | 84.7 | 83.8 |

### 3.2 Classification of the MI EEG signals in BCI Competition IV

In this subsection, MI EEG signals of two and four classes will be examined separately. In this respect, the two-class BCI Competition IV (2008) dataset IIb and four-class BCI Competition IV (2008) dataset IIa will be introduced, and results of the experiments involving transformation and no transformation applied to these databases will be presented.

#### 3.2.1 Classification of two-class MI EEG signals

The BCI Competition IV (2008) dataset IIb contains epochs belonging to two classes collected from nine subjects. Each epoch contains three-channel EEG signals. As done before for the other datasets, 80% of whole data is allocated for the training and 20% for the test. In this IIb dataset, the first two sessions (..01T, ..02T) contain training data without feedback, and the last three sessions (..03T, ..04E, ..05E) with smiley feedback. These no feedback sessions are more challenging to use. Each session belonging to each subject contains 120 epochs. In this study, since the first two sessions (..01T, ..02T) without feedback is used, there are a total of 240 epochs for each subject. The number of epochs in the training set is increased by ten times by the augmentation process. Therefore, for each subject the number of epochs in the training and test sets are 10 ×240 ×0.8 and 240 ×0.2, respectively.

Table 12 shows the classification results of the MI EEG signals processed without using transformation, and presents the comparisons between the proposed DivFE and the studies in the literature in terms of the classification performances. Table 13 presents the FE structures specifically designed for each subject to generate the classification results in Table 12.

Table 14 shows the classification results of the MI EEG signals processed by using transformation (bandpass filters + CSP). Comparative performance results between the proposed DivFE and the studies in the literature are also presented in Table 14. Table 15 presents the FE structures specifically designed for each subject to generate the classification results in Table 15.

**Table 11** The FE structures which generate the classification results in Table 10.

| | #input nodes, filter size, #feature planes in FE layers | | |
|---|---|---|---|
| **FE layers** | **S1** | **S2** | **S3** |
| **Layer 1** | 28, 7, 60 | 24, 15, 40 | 20, 7, 60 |
| **Layer 2** | 60, 7, 60 | 40, 15, 40 | 60, 7, 60 |
| **Layer 3** | 60, 7, 60 | 40, 15, 40 | 60, 7, 60 |
| **Layer 4** | 60, 7, 60 | 40, 15, 40 | 60, 7, 60 |
| **Layer 5** | 60, 40, 16 | 40, 40, 16 | 60, 40, 16 |

### 3.2.2 Classification of four-class MI EEG signals

The BCI Competition IV (2008) dataset IIa contains epochs belonging to four classes from nine subjects. Each epoch contains 22-channel MI EEG signals. There are two sessions for each subject and each session contains 288 epochs. So, there are a total of 576 epochs for each subject. 80% of the data collected from the subjects is allocated for the training and the remaining 20% for the test. The number of epochs in the training set is increased by ten times after the augmentation process. Therefore, the number of epochs in the training and test sets for each subject are $10 \times 576 \times 0.8$ and $576 \times 0.2$, respectively.

**Table 12** Classification results for the two-class MI EEG signals without using transformation.

| | % positive and negative predictive values for each class<br>% accuracies obtained for each subject | | | | | | | | | |
|---|---|---|---|---|---|---|---|---|---|---|
| | **S1** | **S2** | **S3** | **S4** | **S5** | **S6** | **S7** | **S8** | **S9** | Mean % Accuracy (Kappa) |
| ppv for class 1 by DivFE | 100 | 95.9 | 81.0 | 100 | 93.3 | 100 | 97.2 | 78.4 | 73.8 | |
| npv for class 2 by DivFE | 90.9 | 71.6 | 84.8 | 96.1 | 89.7 | 88.5 | 85.6 | 100 | 92.2 | |
| average accuracy of DivFE | 95 | 79.5 | 82.9 | 98 | 91.4 | 93.5 | 90.6 | 86.3 | 80.5 | 88.6 (0.772) |
| Chin's method [22] | 55 | 40.7 | 41.5 | 96.2 | 89.5 | 70.7 | 67 | 88.7 | 80.5 | 69.9 |
| Mengxi's method [9] | 64.1 | 50.9 | 57.7 | 93.1 | 73.4 | 73.1 | 66.6 | 62.9 | 63.8 | 67.3 |
| Yousef's method [9] | 76.0 | 65.8 | 75.3 | 95.3 | 83.0 | 79.5 | 74.5 | 75.3 | 73.3 | 77.6 |

**Table 13** The FE structures which generate the classification results in Table 12.

| FE Layers | #input nodes, filter size, #feature planes in FE layers | | | | | | | | |
|---|---|---|---|---|---|---|---|---|---|
| | **S1** | **S2** | **S3** | **S4** | **S5** | **S6** | **S7** | **S8** | **S9** |
| **Layer 1** | 3,5,60 | 3,9,40 | 3,9,60 | 3,9,40 | 3,9,40 | 3,9,40 | 3,9,40 | 3,9,40 | 3,9,40 |
| **Layer 2** | 60,5,60 | 40,9,40 | 60,9,60 | 40,9,40 | 40,9,40 | 40,9,40 | 40,9,40 | 40,9,40 | 40,9,40 |
| **Layer 3** | 60,5,60 | 40,9,40 | 60,9,60 | 40,9,40 | 40,9,40 | 40,9,40 | 40,9,40 | 40,9,40 | 40,9,40 |
| **Layer 4** | 60,5,60 | 40,9,40 | 60,9,60 | 40,9,40 | 40,9,40 | 40,9,40 | 40,9,40 | 40,9,40 | 40,9,40 |
| **Layer 5** | 60,47,16 | 40,47,16 | 60,9,60 | 40,47,16 | 40,47,16 | 40,47,16 | 40,47,16 | 40,9,40 | 40,9,40 |
| **Layer 6** | | | 60,24,16 | | | | | 40,24,16 | 40,24,16 |

Table 16 shows the classification results of the MI EEG signals processed without using transformation, and the comparisons between the proposed DivFE and the studies in the literature in terms of the classification performances are presented. Table 17 presents the FE structures designed for each subject to generate the classification results in Table 16.

Table 18 shows the classification results of the MI EEG signals processed by using transformation (bandpass filters + CSP), and presents the comparative classification performances of the proposed DivFE method and the studies in the literature. Table 19 presents the FE structures designed for each subject to generate the classification results in Table 18.

**Table 14** Classification results for the two-class MI EEG signals by using transformation.

| | % positive and negative predictive values for each class<br>% accuracies obtained for each subject | | | | | | | | | |
|---|---|---|---|---|---|---|---|---|---|---|
| | **S1** | **S2** | **S3** | **S4** | **S5** | **S6** | **S7** | **S8** | **S9** | Mean % Accuracy (Kappa) |
| ppv for class 1 by DivFE | 97.5 | 69.9 | 67.9 | 95.8 | 94.3 | 83.2 | 84.4 | 77.4 | 92.8 | |
| npv for class 2 by DivFE | 93.3 | 100 | 84.7 | 89.2 | 97.9 | 96.5 | 81.6 | 79.9 | 73.1 | |
| average accuracy of DivFE | 95.3 | 78.5 | 73.7 | 92.3 | 96 | 88.8 | 82.9 | 78.7 | 80 | 85.1 (0.702) |
| Chin's method [22] | 55 | 40.7 | 41.5 | 96.2 | 89.5 | 70.7 | 67 | 88.7 | 80.5 | 69.9 |
| Mengxi's method [9] | 64.1 | 50.9 | 57.7 | 93.1 | 73.4 | 73.1 | 66.6 | 62.9 | 63.8 | 67.3 |
| Yousef's method [10] | 76.0 | 65.8 | 75.3 | 95.3 | 83.0 | 79.5 | 74.5 | 75.3 | 73.3 | 77.6 |

### 3.3 Contribution of the augmentation process to the classification of MI EEG signals

In this subsection, the advantages of the augmentation process will be demonstrated. For the cases when transformation is applied to the raw data, there is a slight difference between the classification performances obtained with augmented and non-augmented training data. But it is observed that if transformation is not used, there is a huge difference between the classification results produced by the augmentation and non-augmentation processes.

**Table 15** The FE structures which generate the classification results in Table 14

| FE Layers | #input nodes, filter size, #feature planes in FE layers | | | | | | | | |
|---|---|---|---|---|---|---|---|---|---|
| | **S1** | **S2** | **S3** | **S4** | **S5** | **S6** | **S7** | **S8** | **S9** |
| **Layer 1** | 8,9,40 | 6,9,40 | 2,17,40 | 12,9,40 | 10,9,40 | 2,9,40 | 2,9,40 | 14,9,40 | 10,9,40 |
| **Layer 2** | 40,9,40 | 40,9,40 | 40,17,40 | 40,9,40 | 40,9,40 | 40,9,40 | 40,9,40 | 40,9,40 | 40,9,40 |
| **Layer 3** | 40,9,40 | 40,9,40 | 40,17,40 | 40,9,40 | 40,9,40 | 40,9,40 | 40,9,40 | 40,9,40 | 40,9,40 |
| **Layer 4** | 40,9,40 | 40,9,40 | 40,17,40 | 40,9,40 | 40,9,40 | 40,9,40 | 40,9,40 | 40,9,40 | 40,9,40 |
| **Layer 5** | 40,47,16 | 40,9,40 | 40,47,16 | 40,47,16 | 40,47,16 | 40,47,16 | 40,47,16 | 40,47,16 | 40,47,16 |
| **Layer 6** | | 40,24,16 | | | | | | | |

Table 20 shows the comparison of classification results obtained using transformation stage (TS) and not using the transformation stage (NTS) with augmented (A) and non-augmented (NA) datasets for two-class problems in the BCI Competition III (2005) dataset IVa which contains 68-channel MI EEGs. Since the channel number is high, the $w$ values of the CSP are determined correctly, thereby achieving high performances without the need for an augmentation process in the case of using transformation. But, in the case of not using the transformation, the classification performance obtained using the augmentation process increases by 22.9 points (in percentage) compared to that obtained without the augmentation process.

**Table 16** Classification results for the four-class MI EEG signals without using transformation.

| | % accuracies obtained for each subject | | | | | | | | | |
|---|---|---|---|---|---|---|---|---|---|---|
| | **S1** | **S2** | **S3** | **S4** | **S5** | **S6** | **S7** | **S8** | **S9** | Mean % Accuracy (Kappa) |
| DivFE for class 1 | 100 | 61.8 | 100 | 81.6 | 41.1 | 32.5 | 81.4 | 100 | 79.8 | |
| DivFE for class 2 | 67.7 | 69 | 100 | 57.1 | 67.2 | 79 | 100 | 83.1 | 88.5 | |
| DivFE for class 3 | 100 | 61.8 | 81.4 | 81.6 | 56 | 74.4 | 82.6 | 89.7 | 87.8 | |
| DivFE for class 4 | 92.7 | 69 | 84.8 | 65.3 | 78.5 | 74.4 | 91.7 | 93.4 | 100 | |
| average accuracy of DivFE | 90 | 65.4 | 91.6 | 71.4 | 60.7 | 65.1 | 88.9 | 91.5 | 89 | 79.3 (0.724) |
| Ang's method [23] | 76 | 56.5 | 81.2 | 61 | 55 | 45 | 82.7 | 81.2 | 70.7 | 67.7 |
| Waseem's method [7] | 77.5 | 58.7 | 82 | 68.5 | 79 | 63.2 | 72.2 | 67 | 70 | 70.9 |
| Raza's method [24] | 91 | 41.5 | 91 | 54.2 | 64.7 | 49.7 | 53.5 | 88.7 | 85.7 | 68.9 |

**Table 17** The FE structures which generate the classification results in Table 16.

| FE Layers | #input nodes, filter size, #feature planes in FE layers | | | | | | | | |
|---|---|---|---|---|---|---|---|---|---|
| | **S1** | **S2** | **S3** | **S4** | **S5** | **S6** | **S7** | **S8** | **S9** |
| **Layer 1** | 22,7,40 | 22,7,60 | 22,7,40 | 22,7,60 | 22,17,40 | 22,7,40 | 22,7,60 | 22,7,60 | 22,7,40 |
| **Layer 2** | 40,7,40 | 60,7,60 | 40,7,40 | 60,7,60 | 40,17,40 | 40,7,40 | 60,7,60 | 60,7,60 | 40,7,40 |
| **Layer 3** | 40,7,40 | 60,7,60 | 40,7,40 | 60,7,60 | 40,17,40 | 40,7,40 | 60,7,60 | 60,7,60 | 40,7,40 |
| **Layer 4** | 40,7,40 | 60,7,60 | 40,7,40 | 60,7,60 | 40,17,40 | 40,7,40 | 60,7,60 | 60,7,60 | 40,7,40 |
| **Layer 5** | 40,29,16 | 60,7,60 | 40,29,16 | 60,29,16 | 40,17,40 | 40,7,40 | 60,29,16 | 60,29,16 | 40,29,16 |
| **Layer 6** | | 60,15,16 | | | 40,15,16 | 40,15,16 | | | |

Table 21 gives a comparison of classification accuracies obtained using transformation stage (TS) and not using the transformation stage (NTS), with augmented (A) and non-augmented (NA) datasets for four-class problems in the BCI Competition III (2005) dataset IIIa which contains 43-channel MI EEGs. As the number of channels decreases, the *w* values of the CSP are determined with lower accuracies. In the case of using transformation, the classification performance obtained using the augmentation process increases by 50 points (in percentage) compared to that obtained without the augmentation process. In the case of not using transformation, the classification performance obtained using the augmentation process increases by 40.8 points (in percentage) compared to that obtained without the augmentation process.

**Table 18** Classification results for the four-class MI EEG signals by using transformation.

| | % accuracies obtained for each subject | | | | | | | | | |
|---|---|---|---|---|---|---|---|---|---|---|
| | **S1** | **S2** | **S3** | **S4** | **S5** | **S6** | **S7** | **S8** | **S9** | Mean % Accuracy (Kappa) |
| DivFE for class 1 | 91.2 | 77.5 | 86.4 | 59.4 | 68.4 | 40.4 | 84.9 | 74.6 | 58.2 | |
| DivFE for class 2 | 77.1 | 75.7 | 97.2 | 63.2 | 93.6 | 53.9 | 99.1 | 76.3 | 85.4 | |
| DivFE for class 3 | 100 | 68.1 | 97.2 | 95 | 54 | 62.9 | 88.4 | 100 | 93.2 | |
| DivFE for class 4 | 89.1 | 71.8 | 90 | 71.3 | 79.2 | 80.9 | 77.8 | 83.6 | 85.4 | |
| average accuracy of DivFE | 89.4 | 73.3 | 92.7 | 72.2 | 73.8 | 59.5 | 87.6 | 83.6 | 80.5 | 79.1 (0.721) |
| Ang's method [23] | 76 | 56.5 | 81.2 | 61 | 55 | 45 | 82.7 | 81.2 | 70.7 | 67.7 |
| Waseem's method [7] | 77.5 | 58.7 | 82 | 68.5 | 79 | 63.2 | 72.2 | 67 | 70 | 70.9 |
| Raza's method [24] | 91 | 41.5 | 91 | 54.2 | 64.7 | 49.7 | 53.5 | 88.7 | 85.7 | 68.9 |

Table 22 gives the comparison of classification results obtained for the combinations of TS and NTS methods with the A and NA datasets for two-class problems in the BCI Competition IV (2008) dataset IIb which contains three-channel MI EEG records. As the number of channels decreases, *w* values of CSP cannot be determined correctly. In the case of using transformation, the classification performance obtained using the augmentation process increases by 14.5 points (in percentage) compared to that obtained without the augmentation process. In the case of not using transformation, the classification performance obtained using the augmentation process increases by 20.1 points (in percentage) compared to that obtained without the augmentation process.

**Table 19** The FE structures which generate the classification results in Table 18.

| FE Layers | #input nodes, filter size, #feature planes in FE layers | | | | | | | | |
|---|---|---|---|---|---|---|---|---|---|
| | **S1** | **S2** | **S3** | **S4** | **S5** | **S6** | **S7** | **S8** | **S9** |
| **Layer 1** | 72,15,40 | 96,13,40 | 44,15,40 | 44,9,40 | 36,7,40 | 16,13,40 | 44,13,40 | 20,15,40 | 40,15,40 |
| **Layer 2** | 40,15,40 | 40,13,40 | 40,15,40 | 40,9,40 | 40,7,40 | 40,13,40 | 40,13,40 | 40,15,40 | 40,15,40 |
| **Layer 3** | 40,15,40 | 40,13,40 | 40,15,40 | 40,9,40 | 40,7,40 | 40,13,40 | 40,13,40 | 40,15,40 | 40,15,40 |
| **Layer 4** | 40,15,40 | 40,13,40 | 40,15,40 | 40,9,40 | 40,7,40 | 40,13,40 | 40,13,40 | 40,15,40 | 40,15,40 |
| **Layer 5** | 40,29,16 | 40,13,40 | 40,29,16 | 40,9,40 | 40,29,16 | 40,13,40 | 40,29,16 | 40,29,16 | 40,15,40 |
| **Layer 6** | | 40,15,16 | | 40,15,16 | | 40,15,16 | | | 40,15,16 |

Table 23 shows the comparison of classification results obtained for the combinations of TS and NTS methods with the A and NA datasets for four-class problems in the BCI Competition IV (2008) dataset IIa which is comprised of 22-channel MI EEG data. As the number of channels decreases, *w* values of the CSP cannot be determined correctly. In the case of using transformation, the classification performance obtained using the augmentation process increases by 33.8 points (in percentage) compared to that obtained without the augmentation process. In the case of not using transformation, the classification performance obtained using the augmentation process increases by 30.2 points (in percentage) compared to that obtained without the augmentation process.

**Table 20** Comparison of classification results obtained using the transformation stage (TS) and not using the transformation stage (NTS), with augmented (A) and non-augmented (NA) datasets for two-class problems in BCI Competition III (2005) dataset IVa.

| Transformation-Augmentation-Dataset | % classification accuracies (Kappa) for the subjects | | | | | Mean Accuracy (Kappa) |
|---|---|---|---|---|---|---|
| | **aa** | **al** | **av** | **aw** | **ay** | |
| **TS-A-III/IVa** | 100 (1.0) | 100 (1.0) | 92.8 (0.86) | 100 (1.0) | 100 (1.0) | 98.5 (0.97) |
| **TS-NA- III/IVa** | 96.4 (0.93) | 96.4 (0.93) | 91.0 (0.82) | 96.4 (0.93) | 96.4 (0.93) | 95.3 (0.906) |
| **NTS-A- III/IVa** | 98.1 (0.96) | 100 (1.0) | 87.2 (0.74) | 98.2 (0.86) | 98.2 (0.86) | 96.3 (0.926) |
| **NTS-NA- III/IVa** | 85.4 (0.71) | 60.0 (0.2) | 65.4 (0.31) | 83.6 (0.67) | 72.7 (0.45) | 73.4 (0.468) |

Tables 20 and 22 show the accuracies and Kappa values obtained for the two classes of dataset

III and IV. F-measures (and sensitivities), corresponding to the accuracies presented in Tables 20 and 22, are calculated and given in Tables 24 and 25, respectively.

**Table 21** Comparison of classification results obtained using the transformation stage (TS) and not using the transformation stage (NTS), with augmented (A) and non-augmented (NA) datasets for four-class problems in BCI Competition III (2005) dataset IIIa.

| Transformation-Augmentation-Dataset | % classification accuracies for the subjects | | | Mean Accuracy (Kappa) |
|---|---|---|---|---|
| | **k3b** | **k6b** | **l1b** | |
| **TS-A-III/IIIa** | 93.4 | 97.1 | 97.0 | 95.8 (0.944) |
| **TS-NA- III/IIIa** | 49.1 | 51.1 | 35.2 | 45.1 (0.268) |
| **NTS-A- III/IIIa** | 98.3 | 94.3 | 96.9 | 96.5 (0.953) |
| **NTS-NA- III/IIIa** | 67.7 | 54.2 | 45.4 | 55.7 (0.410) |

Two sets are created with the classification results obtained from 26 subjects for augmentation and non-augmentation processes in the case of applying no transformation. The paired t-test is calculated to demonstrate the effect of using the augmentation process. The violin plot provided in Fig. 4 allows for a visual investigation of the effect of augmentation process on the accuracy of the classification results. It is clear that, in case of augmentation the distribution of the achieved accuracy results is skewed with a mode value of 98.2%, whereas in case of non-augmentation the distribution is almost symmetric with a mode value of 67.2%. A paired t-test further validated that, the obtained accuracy results with augmentation process is statistically significantly higher (p-val: 2.33e–12) than the results obtained with non-augmentation process.

**Table 22** Comparison of classification results obtained using transformation stage (TS) and not using the transformation stage (NTS), with augmented (A) and non-augmented (NA) datasets for two-class problems in BCI Competition IV (2008) dataset IIb.

| Transformation-Augmentation-Dataset | % classification accuracies (Kappa) for the subjects | | | | | | | | | Mean Accuracy (Kappa) |
|---|---|---|---|---|---|---|---|---|---|---|
| | **S1** | **S2** | **S3** | **S4** | **S5** | **S6** | **S7** | **S8** | **S9** | |
| **TS-A-IV/IIb** | 95.3 (0.91) | 78.5 (0.57) | 73.7 (0.47) | 92.3 (0.85) | 96.0 (0.92) | 88.8 (0.78) | 82.9 (0.66) | 78.7 (0.57) | 80.0 (0.60) | 85.1 (0.702) |
| **TS-NA- IV/IIb** | 62.7 (0.25) | 61.9 (0.24) | 57.7 (0.15) | 82.6 (0.65) | 88.4 (0.77) | 77.1 (0.54) | 70.2 (0.40) | 68.0 (0.36) | 67.5 (0.35) | 70.6 (0.413) |
| **NTS-A- IV/IIb** | 95.0 (0.9) | 79.5 (0.59) | 82.9 (0.66) | 98.0 (0.96) | 91.4 (0.83) | 93.5 (0.87) | 90.6 (0.81) | 86.3 (0.73) | 80.5 (0.61) | 88.6 (0.772) |
| **NTS-NA- IV/IIb** | 67.5 (0.35) | 69.2 (0.38) | 65.7 (0.31) | 87.7 (0.75) | 63.8 (0.28) | 65.6 (0.31) | 60.4 (0.21) | 70.4 (0.41) | 66.6 (0.33) | 68.5 (0.370) |

**Table 23** Comparison of classification results obtained using transformation stage (TS) and not using the transformation stage (NTS), with augmented (A) and non-augmented (NA) datasets for four-class problems in BCI Competition IV (2008) dataset IIa.

| Transformation-Augmentation-Dataset | % classification accuracies for the subjects | | | | | | | | | Mean Accuracy (Kappa) |
|---|---|---|---|---|---|---|---|---|---|---|
| | **S1** | **S2** | **S3** | **S4** | **S5** | **S6** | **S7** | **S8** | **S9** | |
| **TS-A-IV/IIa** | 89.4 | 73.3 | 92.7 | 72.2 | 73.8 | 59.5 | 87.6 | 83.6 | 80.5 | 79.1 (0.721) |
| **TS-NA- IV/IIa** | 35.0 | 37.1 | 50.4 | 46.5 | 42.3 | 40.4 | 38.6 | 46.3 | 70.8 | 45.3 (0.270) |
| **NTS-A- IV/IIa** | 90.0 | 65.4 | 91.6 | 71.4 | 60.7 | 65.1 | 88.9 | 91.5 | 89.0 | 79.3 (0.724) |
| **NTS-NA- IV/IIa** | 60.9 | 34.5 | 67.5 | 41.8 | 36.4 | 44.1 | 41.3 | 59.8 | 56.0 | 49.1 (0.322) |

## 4. DISCUSSION

It is known that the spectrum of MI EEG signals varies from subject to subject [15]. Transformations such as CSP, FFT, STFT and wavelet contain many parameters. In the literature, the best values of the parameters of the transformations are investigated depending on the subject in order to increase the classification performance. In [12], wavelet transforms were applied to the MI EEG signals. Since the wavelet transform is affected by the frequency bands specific to the subject, the use of CSP instead of wavelet transformation was preferred in this and in our previous studies [15−17].

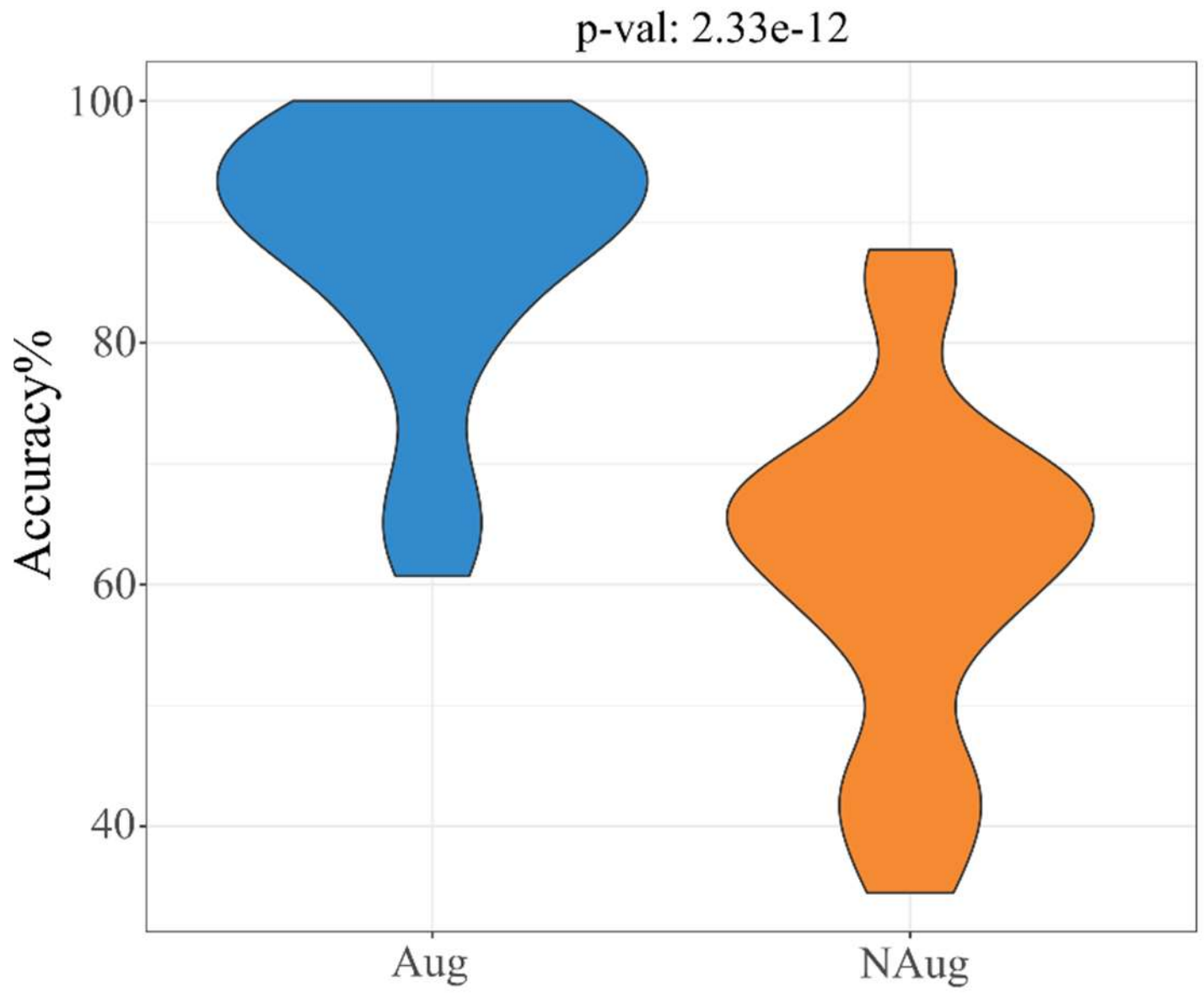

Fig 4. The paired t-test to demonstrate the effect of using the augmentation process.

The AlexNet architecture used in [12] gives 99.35% performance for only two classes of one dataset (BCI III dataset IVa). However, it is natural to achieve such a high classification performance by employing excessive number of weights with a structure of five convolutional layers and two fully

**Table 24** F-measures (and sensitivities) corresponding to the accuracies in Table 20.

| Transformation-Augmentation-Dataset | aa | al | av | aw | ay |
|---|---|---|---|---|---|
| **TS-A-III/IVa** | 1.0 (1.0) | 1.0 (1.0) | 0.920 (0.864) | 1.0 (1.0) | 1.0 (1.0) |
| **TS-NA-III/IVa** | 0.968 (0.964) | 0.966 (0.968) | 0.902 (0.84) | 0.966 (0.964) | 0.962 (0.95) |
| **NTS-A-III/IVa** | 0.984 (1.0) | 1.0 (1.0) | 0.918 (0.856) | 0.983 (1.0) | 0.980 (1.0) |
| **NTS-NA-III/IVa** | 0.875 (0.836) | 0.621 (0.872) | 0.678 (0.8) | 0.836 (0.72) | 0.681 (0.872) |

**Table 25** F-measures (and sensitivities) corresponding to the accuracies in Table 22.

| Transformation-Augmentation-Dataset | S1 | S2 | S3 | S4 | S5 | S6 | S7 | S8 | S9 |
|---|---|---|---|---|---|---|---|---|---|
| **TS-A-IV/IIb** | 0.952 (1.0) | 0.823 (0.715) | 0.773 (0.857) | 0.92 (0.96) | 0.962 (0.983) | 0.895 (0.871) | 0.826 (0.837) | 0.792 (1.0) | 0.765 (0.944) |
| **TS-NA-IV/IIb** | 0.579 (0.604) | 0.652 (0.691) | 0.6 (0.578) | 0.809 (0.807) | 0.909 (0.849) | 0.789 (0.742) | 0.731 (0.668) | 0.706 (0.553) | 0.666 (0.55) |
| **NTS-A-IV/IIb** | 0.944 (0.93) | 0.75 (1.0) | 0.833 (0.896) | 0.978 (0.984) | 0.913 (0.98) | 0.933 (0.971) | 0.9 (0.89) | 0.880 (0.808) | 0.829 (0.75) |
| **NTS-NA-IV/IIb** | 0.666 (0.6) | 0.647 (0.615) | 0.647 (0.685) | 0.893 (0.875) | 0.605 (0.851) | 0.645 (0.562) | 0.541 (0.512) | 0.698 (0.772) | 0.571 (0.555) |

connected layers in the DNN. When we make a layer-wise calculation of the number of weights, it can easily be observed that the excessive increase in the number of weights comes from the fully connected layers. The number of weights contained in the convolutional layers can be calculated as follows:

$3\times11\times11\times96 + 96\times5\times5\times256 + 256\times3\times3\times192 + 192\times3\times3\times192 + 192\times3\times3\times128 = 1.644.576$

The number of weights coming from the fully connected layers is as follows:

$13\times13\times128\times2048 + 2048\times2048 + 2048\times2 = 48.500.736$

The number of weights used in DivFE for subject S1 in Table 7 is given below. The DivFE for subject S1 gives 100% classification performance. There are four convolutional layers and one layer for the minimum distance network in the DivFE. The CSP for subject S1 has only two outputs. The size of the filters is 7 for all layers. After five max-pooling processes, the input data with 251 time samples is reduced to 16 samples. The dimension of the Walsh vectors is 16. The number of weights by using the structure in Table 7 is calculated below:

2×7×40 + 40×7×40 + 40×7×40 + 40×7×40 + 40×16×16 + 16×2= 44.432

Since most of the nodes (weights) in the deep neural network are in the FCNN part, the MDN is preferred rather than the FCNN in this study.

When the studies in Tables 2, 3, 8 and 16 are examined, it is observed that one versus one (OVO) [26, 37] or one versus rest (OVR) [6, 15, 21, 22, 24, 30,] methods are employed to increase the classification performance of MI EEG signals with four classes. In the OVO method, as the number of classes increases, the number of DNNs needed increases excessively. For example, six DNNs should be employed for a four-class problem, and each DNN will be trained individually. The number of DNNs would be 28 for an eight-class problem. In the OVR method, the number of DNNs needed is equal to the number of classes. Because the number of weights in DivFE is low, the use of the OVR method will not pose a problem. Thus, higher performances will be achieved with DivFEs. Let's examine the effects of the OVO and OVR methods on the classification performance of the DivFE with an example. The classification accuracy for subject S5 is obtained as 60.7% in Table 16, and this value is the lowest success rate achieved by the DivFE in Table 16. Tables 26 and 27 show the classification performances and the number of weights of the DivFE with OVO and OVR methods for subject S5 in Table 16, respectively. It is observed that higher performances are obtained by using the OVO and OVR methods. The advantages of OVR method on the performance of DivFE using the four BCI Competition datasets will be investigated in a future study.

In the tables presenting the FE structures, similarities are observed in some network structures that give high classification accuracies. In future studies, a single network structure that provides high performances for all the subjects will be investigated. The training will be guided by the augmentation process so that the DivFE will give the best performances. In this sense, a new augmentation process will be investigated for a single network structure.

**Table 26** The classification performances and the number of weights of DivFE with OVO method.

| | **class 1 vs 2** | **class 1 vs 3** | **class 1 vs 4** | **class 2 vs 3** | **class 2 vs 4** | **class 3 vs 4** |
|---|---|---|---|---|---|---|
| **Accuracy** | 0.865 | 0.796 | 0.811 | 0.811 | 0.811 | 0.815 |
| **Number of weights** | 58320 | 58320 | 58320 | 58320 | 58320 | 58320 |

**Table 27** The classification performances and the number of weights of DivFE with OVR method.

| | **class 1** | **class 3** | **class 3** | **class 4** |
|---|---|---|---|---|
| **Accuracy** | 0.906 | 0.849 | 0.839 | 0.887 |
| **Number of weights** | 58320 | 58320 | 58320 | 58320 |

As can be observed from Tables 22−25, we have demonstrated that by resulting in high success rates for the classification of MI EEGs, the augmentation process is able to compete with the CSP. The obtained results show that there is no need to use CSP, augmentation alone is sufficient for high successes. The augmentation process should be applied after the data is separated into training, validation and test sets. 80% of the small-sized original dataset should be reserved for the training set and 20% for the test set. After allocating 10% of the training set for validation, the remaining data in the training set (which is 72% of whole dataset) is augmented. Hence, augmentation process is not applied to validation and test data. The performance results in all the tables were obtained in this way. If the small-sized original dataset is first augmented and then it is separated into 72% training, 8% validation and 20% test sets, this would not be the correct sequence for the dataset preparation. Because the augmented variants of the original data would exist in the three sets of training, validation and test, DNN would have all the information about the data. In this case, 100% success rates would be obtained for all of the training, validation and test sets of the BCI Competition datasets.

In [40], several training sets were created by using 10%, 20%, 30%, 40% and 50% of whole data. The classification performances were investigated by using the CSP and neural networks together for each training set. In those analyses, the numbers of data in the test sets were kept constant. Actually, since the dataset size was not synthetically increased, the numbers of data in the training sets were low. It has been shown that as the numbers of data in the training sets were increased, the classification performances were improved. Since CSP transformation was used in all the analyses, its contribution on the performance became about 5%. In our study, since the CSP transformation is not used in the classification process, the positive effect of the augmentation process on the performance is found to be approximately 30%.

## 5. CONCLUSIONS

In our previous studies [15−17], MI EEG signals were classified by using a novel CSP method and the LDA classifier without using a DNN. It was observed that classification successes of the CSP based models did not exceed a certain level especially when the number of classes was more than two, and the performances of CSP models were very dependent on the subjects. Since the classification performance generally decreases as the number of classes increases, the correct determination of the features becomes extremely important in increasing the classification performance. In this context, the DNN plays a very important role in the feature extraction stage. It is observed in the literature that high performances are obtained by using the DNNs for MI EEG signals which have more than two classes [6,11,13].

In the classification of MI EEG signals, both methods (with and without transformation stages) were compared in their best conditions. Therefore, optimum $m$ parameters of the CSPs were investigated to increase the classification performance. Epochs are formed by selecting specific time intervals on the EEG recordings. The selection of the beginning and the duration of the epochs have a strong impact on the classification performance [15, 30]. In the study, these values are fixed for all subjects and they are not changed. Only the *m* value is different for all the subjects. These three values (the beginning and the duration of the epochs, and parameter *m*) are determined to give high classification performances which in turn make the classification system dependent on the parameter [30]. When *w* values of CSPs are determined by using the whole dataset (training + test data) of each subject, and CNNs are trained with only the training data (without using the test data), 100% classification performance is obtained for each subject. To obtain the classification results presented in the tables, the *w* weight values of CSPs belonging to each subject were determined using only the training data. In this case, the classification performances were observed to be low. This indicates that the *w* values are highly dependent on the EEG dataset.

When filter bank and CSP were both applied to the data for the transformation, it is observed that the augmentation process had caused a slight increase in the classification performance. However, in the lack of transformation, it is observed that the augmentation process has increased the classification performance about 30%. These two experiments reveal the positive contribution of the augmentation process on the classification performance.

When the proposed methods are compared with those in the field, a slight increase in the classification performance is achieved. It should be noted that, these successes are achieved with a simple network structure and training strategy. Augmenting the dataset has a very positive impact on the FE's training. However, in this study, the only purpose of the augmentation process is not to increase the number of epochs in the dataset, a guidance is also involved in the constitution of the training set in order to prevent overfitting. It is desirable that the training of FE is not affected by instantaneous amplitude changes on EEG signals and occurrence of motor imagery at different starting points within the epoch. Therefore, the amplitudes of the data in the training set were randomly changed and the signal was rotated in the epoch.

Low performances as reported in the literature have been obtained for some subjects in the classification of MI signals of BCI Competition IV dataset IIa. It is a common problem that some subjects cannot properly participate in the MI based experiments, hence low performances are obtained with those subjects.

In deep learning applications, one of the main problems is to determine the coarse structure of DNN. With a small-size DNN structure, it is easier to search for a coarse model. Since there are a few parameters in DivFE, the number of layers and the number of features in these layers are easily determined leading to fast training of the DivFE. Moreover, the problems caused by the fully connected network were not dealt with during the training phase.

The OVO and OVR methods have the potential of improving the classification performance of MI EEG signals. It is observed that higher classification performances for multi-class problems were achieved by using the OVO [26, 37] and OVR [6, 15, 21, 22, 24, 30] methods. OVO and OVR methods do not only produce advantages when CSP is used, but also high performance is achieved by DNNs without any transformations, as we have demonstrated in the *Discussion* section. In [26], the average classification accuracy for subject S5 is achieved as 79.2% with the OVO method. In our study, provided in Table 16, the average classification accuracies for subject S5 increased from 60.7% to 81.8% and 87.0% with the OVO and OVR methods, respectively. We think that the differences between the results obtained with the OVO (81.8%) and OVR (87.0%) are due to the training strengthened with the augmentation process. Actually, the proposed augmentation process can be used alone with any DNN model to increase the classification performance of MI EEG signals.

**Acknowledgement**

This work was supported by the Istanbul Technical University Scientific Research Project Unit [ITU-BAP MYL-2018-41621].